\documentclass[lettersize,journal]{cas-dc}
\usepackage{amsmath,amsfonts}
\usepackage[authoryear]{natbib}
\usepackage{algorithmic}
\usepackage{adjustbox}
\usepackage{listings}
\usepackage{algorithm}
\usepackage{array}
\usepackage{subcaption}
\usepackage{textcomp}
\usepackage{stfloats}
\usepackage{url}
\usepackage{verbatim}
\usepackage{fvextra}
\usepackage{graphicx}
\usepackage{enumitem}
\usepackage{ragged2e}
\usepackage{gensymb}
\usepackage{cite}
\usepackage{tabularx}
\usepackage{capt-of}
\usepackage{xcolor}
\usepackage{tcolorbox}
\usepackage{tikz}
\usetikzlibrary{arrows.meta, positioning, shadows, shapes.geometric, shapes.multipart}

\tikzstyle{startstop} = [rectangle, rounded corners, minimum width=3cm, minimum height=1.5cm,text centered, draw=black, fill=red!30]

\tikzstyle{process} = [rectangle, minimum width=3.5cm, minimum height=1cm, text centered, draw=black, fill=blue!30]
\tikzstyle{arrow} = [thick,->,>=stealth]

\lstdefinelanguage{json}{
    basicstyle=\ttfamily\footnotesize,
    breaklines=true,
    numbers=none,
    stepnumber=1,
    showstringspaces=false,
    tabsize=2
}

\lstdefinestyle{textlist}{
    basicstyle=\ttfamily\scriptsize,
    breaklines=true,
    numbers=none,
    showstringspaces=false
}
\newcounter{numquote}
    {\par} 

\begin{document}
\let\WriteBookmarks\relax
\def\floatpagepagefraction{1}
\def\textpagefraction{.001}

\title[mode = title]{Gaussian Building Mesh (GBM): Extract a Building's 3D Mesh with Google Earth and Gaussian Splatting}
\shorttitle{Gaussian Building Mesh}
\shortauthors{K. Gao et al.}  
\author{Kyle Gao}[orcid=0000-0002-8320-6308]
            \credit{Conceptualization, Methodology, Validation, Formal analysis, Investigation, Data Curation, Writing - Original Draft, Visualization}
\ead{y56gao@uwaterloo.ca}

\author{Liangzhi Li}
\author{Hongjie He}
            
\author{Dening Lu}
\credit{Investigation, Writing - Review \& Editing}

\author{Linlin Xu}
            \credit{Funding Acquisition, Resources, Writing - Review \& Editing, Supervision}

\author{Jonathan Li}
\credit{Funding Acquisition, Resources, Writing - Review \& Editing, Supervision}
\ead{junli@uwaterloo.ca}


\maketitle
\begin{abstract}
The rapid convergence of computer vision, and digital technologies is redefining how buildings are captured, modeled, and managed. In computer vision, Recently released open-source pre-trained foundational image segmentation and object detection models allow for geometrically consistent segmentation of objects of interest in multi-view 2D images. Text-based or click-based prompts can be used to segment objects of interest without requiring labeled training datasets, allowing for both user-prompted or automated segmentation. Simultaneously, Gaussian Splatting allows for learning a 3D representation of a scene's geometry and radiance based on 2D images. Combining Google Earth Studio, SAM2+GroundingDINO, 2D Gaussian Splatting, and our improvements in mask refinement based on morphological operations and contour simplification, we created a pipeline to extract the 3D mesh of any building based on its name, address, or geographic coordinates. Our pipeline offers a fast and user-accessible for rapid 3D modeling of built environments and structures, enabling downstream applications.   
\end{abstract}



\section{Introduction}
The extraction of 3D building models from remote sensing images has long been an active research topic, with applications ranging from urban planning, disaster management, environmental monitoring, telecommunications, construction, digital media, and many more. In remote sensing, the standard way to extract 3D information from 2D images is by using photogrammetry, which involves identifying key points in multi-view images of a scene and then triangulating and registering these key points into a cohesive 3D point cloud of the scene. Recent innovations in learning-based 3D rendering approaches, namely Gaussian Splatting, have opened up new possibilities in learning both accurate 3D lighting and 3D geometry from 2D images, attracting much research interest. Additionally, advances in deep learning image processing have greatly improved the capabilities of extracting individual objects from images.

Leveraging Google Earth Studio \citep{google_earth_studio}, and inspired by GS2Mesh\citep{2024gs2mesh}, we propose a 3D building mesh extraction pipeline capable of extracting the 3D mesh of a building given its proper name, address, postal code, or geographical coordinates. When combined with off-the-shelf registration methods, our pipeline enables downstream tasks such as building information models, construction verification, and automated safety assessment.

Our contributions are as follows:
\begin{itemize}
    \item We create a novel meshing pipeline that allows for the extraction of a 3D mesh of a building from its name, address, postal code, or latitude/longitude coordinates without using on-site LiDAR, or camera data.
    \item We leverage and improve Segment Anything Model-2 \citep{SAM2} and GroundingDINO \citep{groundingdino} for geometrically consistent building masking. We add mask refinements based on morphological operations and the Ramer-Douglas-Peucker algorithm combined with mask re-prompting, which can aid in other use cases of SAM2 and GroundingDINO.
    \item We improved and modified an unpublished implementation of 2DGS \citep{20242dgs} to generate 3D colored building meshes via masked TSDF integration, with refinements in depth map filtering, smoothing, and hyperparameter tuning during both training and meshing.
\end{itemize}

GS2Mesh \citep{2024gs2mesh} combines out-of-the-box object segmentation and 3DGS to generate 3D meshes of objects regardless of the background. To our knowledge, it is the only method similar to our own that performs object-based 3D mesh extraction from text prompts or click prompts. This method was the main source of our inspiration. There are many other Gaussian Splatting-based 3D geometry and mesh extraction methods. However, they focus on reconstructing the entire scene as opposed to being able to extract the mesh of designated objects based on user-based text input, clicks, or system-level automation.

\vspace{0.5cm}
\begin{tikzpicture}[node distance=1.5cm, scale=0.75, transform shape]
\begin{scope}[shift={(100pt,0)}]

\node (start) [startstop] {Google Earth Studio};
\node[above=0.4cm of start] (input) {One of \{\textit{address, place name, postal code,  geographic coordinates}\}};
\node (process1) [process, below of=start] {SAM2 Mask Extraction};
\node (process2) [process, below of=process1] {Mask Refinement};
\node (process3) [process, below of=process2] {2D Gaussian Splatting};
\node (process4) [process, below of=process3] {TSDF Fusion};

\draw [arrow] (input) --  (start);
\draw [arrow] (start) --  (process1);
\draw [arrow] (process1) -- (process2);
\draw [arrow] (process2) -- (process3);
\draw [arrow] (process3) -- (process4);
\draw [arrow, bend left] (start.east) to node[midway, right] {Remote sensing images} (process1.north east);
\draw [arrow, bend left] (process1.east) to node[midway, right] {Images and masks} (process2.north east);
\draw [arrow, bend left] (process2.east) to node[midway, right] {Images and refined masks} (process3.north east);
\draw [arrow, bend left] (process3.east) to node[midway, right] {Depth and color maps} (process4.north east);

\draw [arrow] (process3.east) -- ++(1cm,0) node[right] {Synthesized images};
\draw [arrow] (process4.east) -- ++(1cm,0) node[right] {3D colored mesh};

\end{scope}
\end{tikzpicture}

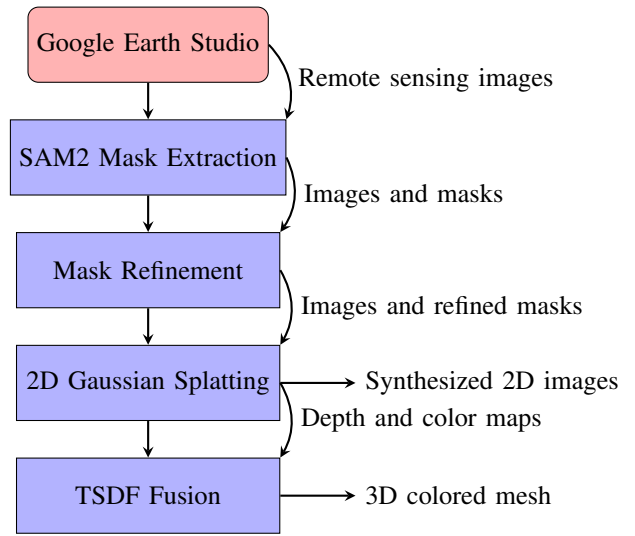
\captionof{figure}{Flow chart of our pipeline. The output data modality at each step is denoted on the right. Processes and modules are boxed in blue.} \label{fig:pipeline}

\vspace{0.5cm}

\section{Related Works}
\subsection{3D Reconstruction from 2D Images}
Classical 3D reconstruction from 2D images is a well-studied area. Photogrammetry is well-understood in remote sensing and computer vision and is widely used for academic, industrial, and commercial purposes. For this purpose, multiple software suites and code libraries \citep{2016COLMAP, bundler, opensfm, realitycapture, envi, arcgis} are commercially available or open-sourced, with some specifically designed for remote sensing applications. Many of these not only allow for the extraction of a sparse point cloud from 2D images but also generate a 3D mesh. These 3D meshes are often more desirable than point clouds, as they are more photorealistic and allow for use in 3D modeling and simulations. COLMAP \citep{2016COLMAP}, in particular, is worth highlighting. It is a photogrammetry and Structure-from-Motion (SfM) library widely used as a preprocessing step for modern novel view synthesis and learning-based 3D rendering methods and is included in most Gaussian Splatting models' pipelines, including our own.

\subsection{Modern Novel-View Synthesis and Neural/Learning-based Rendering}
We categorize two broad families of recently developed methods in this section: NeRF and Gaussian Splatting, with a focus on papers that apply these techniques to remote sensing and building model extraction. Given the extensive body of work in this field, it is more comprehensively captured in survey papers \citep{gaosurvey, 3dgssurvey} than in a few paragraphs of literature review.

Both NeRF and Gaussian Splatting methods are fundamentally based on two core concepts: differentiable rendering/rasterization and learning-based 3D representation. Differentiable rendering and rasterization enable the computation of gradients during the creation of 2D images from a 3D representation. This, in turn, allows for the learning and refinement of the 3D representation through loss-function optimization, similar to neural network training.

Neural Radiance Field (NeRF) \citep{nerf}, introduced in 2020, has garnered significant attention in this field. In NeRF-based models, the 3D representation consists of a 3D radiance field (directionally dependent color field) and a 3D density field, both represented as Multi-Layer Perceptrons (MLPs). Differentiable volume rendering is employed to generate 2D images by sampling and integrating local 3D color/radiance and density. These radiance and density fields are trained from scratch using a photometric loss function, meaning NeRF models learn a 3D representation of a scene from 2D images. Significant advancements have been made in this area, including \citep{mipnerf, mipnerf360, instantngp}, with models like \citep{neus, unisurf, neus2} focusing on improving 3D geometry extraction, and others like \citep{BungeeNeRF, urbanrf, blocknerf, shadownerf, satnerf} applying NeRF techniques to remote sensing or urban scene capture.

3D Gaussian Splatting (3DGS) \citep{2023gaussian_splatting}, first proposed in 2023, and subsequent Gaussian Splatting models have largely surpassed NeRF-based approaches over the past year. 3DGS uses a large number of 3D Gaussian distributions (also known as primitives in computer graphics) as a learned 3D representation. In addition to standard 3D Gaussian distribution function parameters, each Gaussian primitive also has directionally dependent color and opacity $\alpha$, which are all trainable parameters. The method uses a differentiable tile-based rasterizer, projecting the 3D Gaussians onto the to-be-rasterized image, and $\alpha$-blending the projected Gaussians. This allows the 3D Gaussians representing the scene's radiance and geometry to be learned from scratch using a photometric loss function. Compared to NeRF, Gaussian Splatting models are generally much faster to train, have higher view-synthesis quality, but require more memory. Methods such as \citep{20242dgs, 2024sugar, 2024GOF} improved 3D extraction, and methods such as \citep{3dgsurban,3dgsurban2,3dgsremote,3dgsremote2,3dgsremote3} applied Gaussian Splatting to remote sensing or urban scene capture.

However, we note that these methods all focus on scene-wide capture and cannot extract the 3D mesh of individual objects without further processing. Only GS2Mesh \citep{2024gs2mesh}, being an out-of-the-box pipeline leveraging Segment Anything Model object masking, allows for the extraction of a 3D mesh from user prompting without further training or processing. Unfortunately, according to our preliminary testing, GS2Mesh often fails in remote sensing building extraction scenarios due to issues in the mask generation module and the pretrained deep learning-based stereo depth map module, which motivated our research.

\section{Background}
\subsection{3D Gaussian Splatting}
3D Gaussian Splatting (3DGS) \citep{2023gaussian_splatting} is a view-synthesis technique that enables the learning of a 3D scene's geometry and lighting from multi-view 2D images, which can then be used to rasterize the scene from novel viewpoints. The process begins with an initial point cloud, often generated using COLMAP \citep{2016COLMAP} Structure-from-Motion (SfM). For each point in this cloud, 3DGS initializes a Gaussian primitive that encodes learnable parameters such as mean, covariance, opacity, and local lighting in the red, green, and blue channels, represented as spherical harmonic coefficients.

To render an image of the scene, a differentiable tile-based rasterizer is employed, projecting the Gaussian primitives into 2D on the image plane. These projected Gaussians are alpha-blended to generate the final image. During training, the learning of the Gaussian Splatting parameters is guided by comparing the rasterized image for a given camera pose to the ground truth training image. The difference in pixel values and overall image quality is used to optimize the parameters of the Gaussians.

\subsection{2D Gaussian Splatting}
\begin{figure}[h]
\centering
\includegraphics[width=0.5\textwidth]{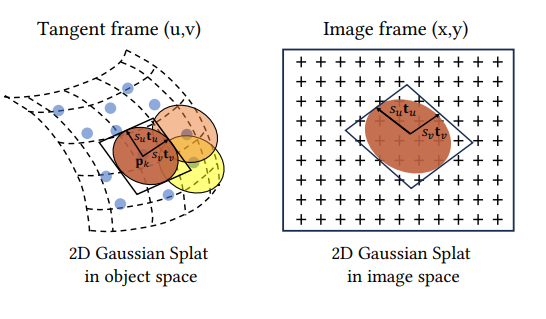}
\caption{2D Gaussian Splatting uses surfaced aligned 2D Gaussian primitives embedded in 3D to represent the 3D scene. 2D Gaussian is represented by it's 3D position $\mathbf{p_k}$, it's scale $s_u, s_v$, and it's orientation $\mathbf{t_u, t_v}$.  \citep{20242dgs}}\label{img:2dgs}
\end{figure}
2D Gaussian Splatting (2DGS) \citep{20242dgs} enhances the standard 3D Gaussian Splatting (3DGS), improving the reproduction of 3D surface geometry. While GS2Mesh slightly outperforms 2DGS on the DTU benchmark, the two methods are fundamentally different. GS2Mesh is primarily a mesh extraction pipeline that utilizes vanilla 3DGS during the 3D reconstruction phase. In contrast, 2DGS is a significant improvement on 3DGS itself, altering the nature of the Gaussian splats. As a result, 2DGS can be used to replace 3DGS in many pipelines, as we have done.

The key innovation in 2D Gaussian Splatting is the representation of the scene using 2D-oriented planar Gaussians instead of 3D Gaussians. Like standard 3DGS, 2DGS employs Gaussian primitives that store spherical harmonic coefficients for each color channel, local transparency $\alpha$, and 3D location $\mathbf{p}_k$. However, unlike 3D Gaussian primitives, 2D Gaussian primitives have two scalar values to represent variance ($s_u, s_v$) and two tangent vectors ($\mathbf{t}_u, \mathbf{t}_v$) whose cross product results in the normal vector that defines orientation (see figure \ref{img:2dgs}). Depth maps can be accurately rendered using the projected depth value. For more details, we refer readers to the original paper \citep{20242dgs}.

\subsection{Segment Anything Model (version 2) and Grounding DINO}
Segment Anything Model (SAM) \citep{SAM} is an out-of-the-box image segmentation model pretrained on a massive billion-image dataset. It is capable of segmenting most objects given point-based or bounding box priors, without requiring further training from the user. Users can provide point-click or bounding box prompts to identify the object(s) of interest, and SAM will return segmentation masks and associated scores. However, a key limitation of SAM is that, when applied to a scene viewed from multiple images, the individual masks generated are not necessarily consistent with each other. This inconsistency limits SAM’s effectiveness in segmenting video data, which requires temporal consistency, and multi-view data, which requires 3D consistency.

SAM2 \citep{SAM2}, released in August 2024, addresses this issue by introducing consistent video segmentation that maintains 3D and temporal consistency through the use of memory attention. Since we aim to extract the mesh of an individual building rather than an entire neighborhood, segmentation masks are crucial.

GroundingDINO \citep{groundingdino} is a pretrained open-set object detector capable of extracting object bounding boxes in images from natural language prompts without requiring additional training from the user. GroundingDINO can be combined with SAM/SAM2 to enable text-based object segmentation. The process involves first generating a bounding box from the text description and then using the bounding box to prompt SAM/SAM2. This combination of GroundingDINO and SAM/SAM2 is known as the Grounded-SAM pipeline \citep{groundedsam}, which is available as an open-source library.

\subsection{GS2Mesh}
GS2Mesh \citep{2024gs2mesh} is a Gaussian Splatting-based 3D reconstruction pipeline, outperforming concurrent and competing methods such as SuGAR \citep{2024sugar}, 2DGS \citep{20242dgs}, and GOF \citep{2024GOF} on the DTU dataset \citep{2016dtu} benchmark. 
\begin{itemize}
    \item GS2Mesh learns and stores the scene in a standard 3DGS model.
    \item The trained 3DGS model is then used to generate a stereo pair for each input image. Each stereo pair is used to generate a depth image.
    \item Grounded-SAM is used to generate multi-view masks to mask out the background for mesh extraction.
    \item A pre-trained depth from stereo model DLNR is used to generate depth maps for each stereo pair. 
    \item  The entire ensemble of depth images are integrated into a mesh using the Truncated Signed Distance Function fusion (TSDF) algorithm \citep{1996tsdf} with the Marching-Cubes algorithm \citep{1998marchingcubes}.
\end{itemize}

More specifically, a standard 3DGS model is trained from the input images. The 3DGS model is then used to generate a stereo pairs for each training image's camera pose. For each pair, the left-image is generated with the same camera pose as the training image, and the right-image is generated with a small shift $[b,0,0]$ to the right. Since Gaussian Splatting models performs best near training poses, this method ensures visual high quality in the generated stereo image pair.

For these stereo image pairs, the Segment Anything Model 2 (SAM2) is used to generate segmentation masks for the objects. In the GS2Mesh paper, which was published before SAM2, the authors addressed the 3D consistency issue by projecting the initial mask onto other frames, sampling new points within the projected mask as SAM prompts, and creating a new SAM mask from these prompts for each frame. The GS2Mesh codebase has since been updated to use SAM2 for 3D geometry-consistent mask generation.

From these stereo pairs, DLNR \citep{2023DLNR}, pretrained on the Middlebury dataset \citep{2002middlefury}, is used for depth extraction from stereo images. To improve the quality of the reconstructions, multiple masks are applied to the stereo model's output to filter out regions visible to only one camera and to discard depth estimates outside the valid range.

Following depth extraction, a standard Open3D \citep{2018open3d} implementation of the TSDF algorithm initializes and populates a voxel grid with the scene geometry. The voxel representation of the scene is populated with the signed distance to the nearest scene surface, integrated from the depth images generated by DLNR. The marching cubes algorithm then assigns each cube’s vertices in the voxel grid to be inside or outside the nearest surface based on the previously calculated TSDF values. Based on the 8-vertex configuration, a local surface is meshed for each cube, which is repeated across the entire voxel grid. This process generates a mesh from the voxel representation.

\subsection{Evaluation Metrics}
For 2D view synthesis visual quality assessment, we use the commonly accepted Peak Signal-to-Noise Ratio (PSNR) \citep{psnr}, 2D Structural Similarity Index Measure (SSIM) \citep{2004ssim}, and Learned Perceptual Image Patch Similarity (LPIPS) \citep{lpips}. These are full-reference metrics that compare an assessed image with a ground truth image. PSNR and SSIM are higher when the assessed image and the ground truth are similar. SSIM achieves a maximum value of 1 when the two images are identical. LPIPS, on the other hand, is lower when the two images are similar, with a minimum of 0 when the two images are identical.

For 3D mesh quality assessment, we use the 3D Structural Similarity Index Measure (3D-SSIM) \citep{3dssim}, comparing a 360-degree rendering video of the mesh with a 360-degree ground truth video created by segmenting the building from its background in the Google Earth training images. We note that there are other full-reference 3D geometry and visual quality metrics that compare 3D models to other 3D models. However, we lack ground truth 3D models for the buildings we meshed and only have access to ground truth 2D images. This is the main reason for using the video-based 3D-SSIM for mesh quality comparison. We provide both the average 3D-SSIM across the entire video and the minimum 3D-SSIM across video frames.

\subsection{Google Earth Studio}
Google Earth Studio \citep{google_earth_studio} is a web-based animation tool. With access to Google's vast collection of 2D and 3D Earth data, ranging from large geological formations to individual buildings, Google Earth Studio provides a simple and efficient way to collect off-nadir images for the training of 3DGS models \citep{gao_3dgs}. Google Earth Studio allows for the specification of a target of interest in terms of longitude and latitude coordinates, address, postal code, or location name. After selecting the target of interest, Google Earth Studio enables the specification of a camera path for which images of the target location are rendered. By selecting a circular camera path orbiting above and pointing towards the target of interest, Google Earth Studio allows for the extraction of a multi-view dataset of the building with 360-degree coverage, well suited for 3D reconstruction.
\section{Methodology} \label{sec:methodology}

Leveraging Google Earth Studio and inspired by GS2Mesh, we created a 3D building mesh extraction pipeline capable of extracting the 3D mesh of a building given its location name, address, postal code, or geographic coordinates. As shown in Figure \ref{fig:pipeline}, our pipeline consists of the following steps:

\begin{enumerate} 
\item \textbf{Multi-view remote-sensing image collection}: We leverage Google Earth Studio to collect multi-view images of a building of interest using its name, address, or postal code or coordinates. 
\item \textbf{Automated building mask extraction}: We automatically extract multi-view consistent building masks for the building of interest in each image using system-level prompts. 
\item \textbf{Building mask refinement}: We refine SAM2 masks using morphological dilation, which slightly extends the mask outwards and fills in holes. We then simplify mask contours using the Ramer-Peuker-Douglas algorithm. Optionally, we also allow for user-based re-prompting to correct possible errors from the automated process. 
\item \textbf{Gaussian Splatting}: We train a 2DGS+ \citep{2dgsp} model to learn building geometry and radiance. 
\item \textbf{Mesh extraction}: We perform masked Truncated Signed Distance Function Fusion on smoothed depth maps to extract a 3D building mesh. We perform mesh refinement and mesh simplification using Open3D triangular mesh post-processing functionalities. \end{enumerate}

\subsection{Google Earth Studio Dataset}
We leverage Google Earth Studio to extract 7 scenes with buildings of interest that we wish to mesh. For each scene, we extract 31 frames/images over a 360$\degree$ circular camera path centered around the building of interest. As part of Google Earth's functionality, we tested and used a variety of methods to identify the buildings of interest, including address, postal code, geographic coordinates, and building name. Each scene's camera information is provided in Table \ref{tab:camera_data}. We rounded the camera tilt to the nearest half-degree. A camera tilt of 0$\degree$ indicates the camera pointing straight down towards the ground, whereas a camera tilt of 90$\degree$ indicates the camera pointing horizontally, parallel to the ground.

\subsection{Mask Extraction and Mesh Refinement} \label{subsection:mask_refinement}
SAM2 provides a rough segmentation mask of the building. This process can be automated by providing system-level prompts to identify the building of interest. We noticed the following problems while inspecting the SAM2 segmentation masks and the 3D mesh extraction that resulted from directly using those masks.

\begin{enumerate} \item \textbf{Poor geometry and holes}: We noticed that SAM2 had trouble with some buildings whose roof or wall colors might be confused with the background, resulting in incomplete masks with holes. 
\item \textbf{Noise pixels and false positives}: SAM2 masks sometimes produced small false-positive pixels away from the object of interest. This occurred at times with multiple identical buildings in close proximity to each other. 
\item \textbf{Poor mask boundary at building base}: SAM2 produced masks with poor geometry at the buildings' base, often resulting in incomplete masking and jagged mask boundaries. This resulted in 3D reconstructions with jagged bottoms, greatly affecting visual quality. 
\item \textbf{Wrong building identification}: When using automated system prompts in scenes where the building of interest is not a key landmark, (E.g. when multiple identical buildings are together), SAM2 at times segmented the wrong building. 
\item \textbf{Incomplete geometry}: When using automated system-level prompting, at times, not the entire building of interest was segmented. Moreover, for complex buildings, SAM2 would at times lose part of the building in long image sequences.
\end{enumerate}

To address these issues, we developed a mask refinement pipeline. We optionally morphologically erode the mask first, removing noise. Then, we grow the mask using morphological dilation to fill in any holes and expand incomplete masks. We then extract the mask's contour using Canny's algorithm and straighten/simplify the contour using the Ramer–Douglas–Peucker algorithm. The contour is then filled into a refined mask, addressing these issues. Our SAM2 image sequence-based building segmentation (mask generation) process is as follows. 

\subsubsection{Automated Mask Extraction} By using the Grounded-SAM pipeline with SAM2, we extract object masks without further training with natural language text. We use the system prompt "the central landmark or building of interest" to automatically extract the multi-view segmentation masks. Additionally, if a uniquely identifying proper noun (e.g. "The Parliament of Canada") was used to extract the multi-view images of the building from Google Earth Studio, we also automatically pass this to the Grounded-SAM pipeline as an additional system prompt. 
\subsubsection{User Re-prompting}
To solve the aforementioned issue of incomplete geometry, we modified the masking pipeline to add optional user re-prompting. We allow users to identify frames with errors and re-prompt the image at those frames. SAM2 re-propagates the user's new prompt throughout all frames, effectively solving these mask inconsistencies.
\begin{table}[htpb]
\caption{Camera altitude and tilt for Google Earth Studio scenes' camera flight path}
\centering
\begin{tabularx}{0.45\textwidth}{l|c|r}
\hline
Location                & Alt. (m) & Tilt ($\degree$) \\ \hline
ICON                    & 680      & 63.5             \\
Canada Parliament       & 119      & 63.5             \\
CN Tower                & 1870     & 47.5             \\
Laurel Heights          & 713      & 57.0             \\
Perimeter Institute     & 882      & 25.5             \\
Dana Porter Library     & 526      & 57.0             \\
Townhouse               & 527      & 45.0             \\ \hline
\end{tabularx}
\label{tab:camera_data}
\end{table}

\begin{figure}[htbp]
    \begin{subfigure}[t]{0.23\textwidth} 
        \centering
        \includegraphics[width=\textwidth]{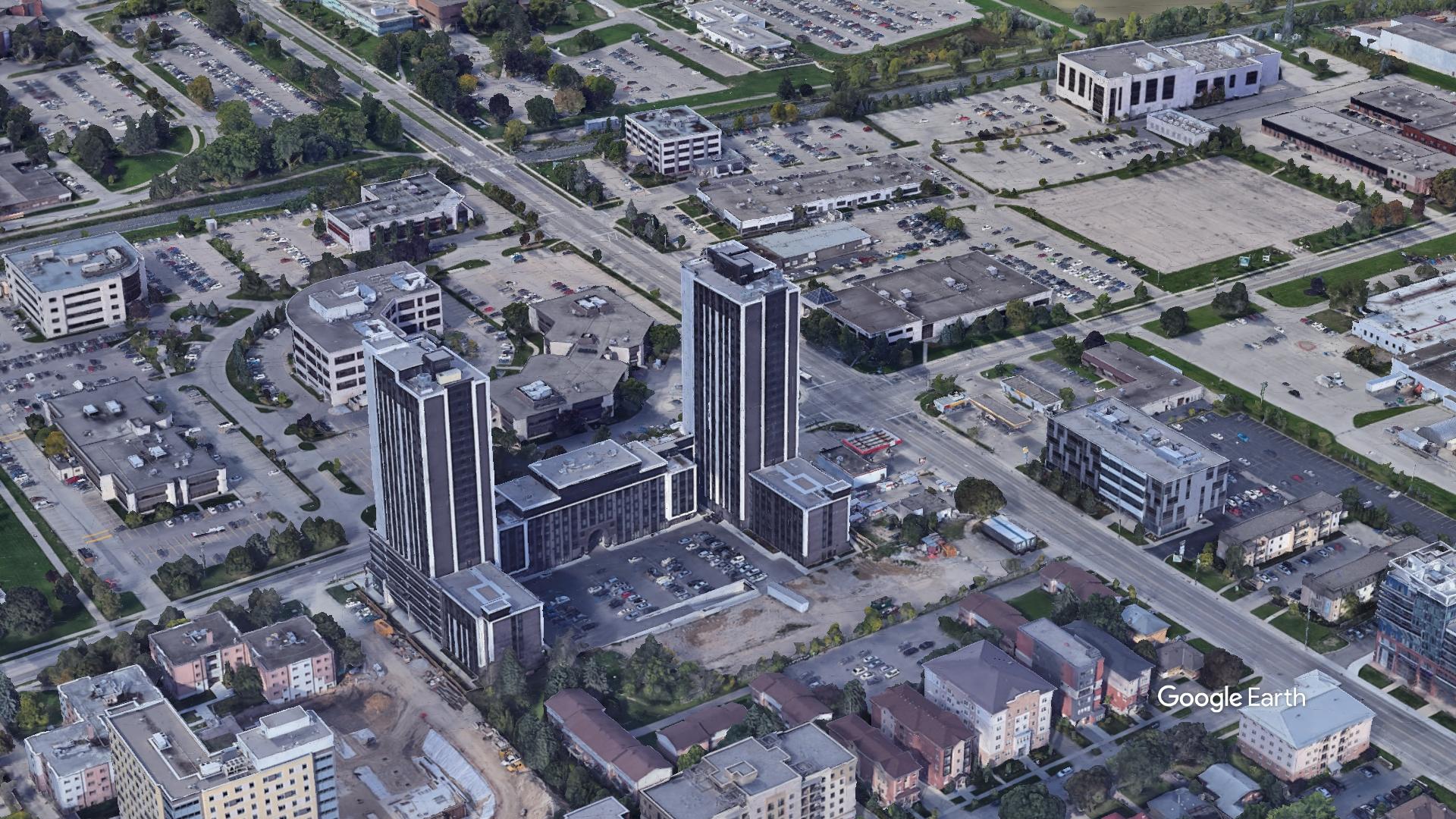}
    \end{subfigure}
    \begin{subfigure}[t]{0.23\textwidth} 
        \centering
        \includegraphics[width=\textwidth]{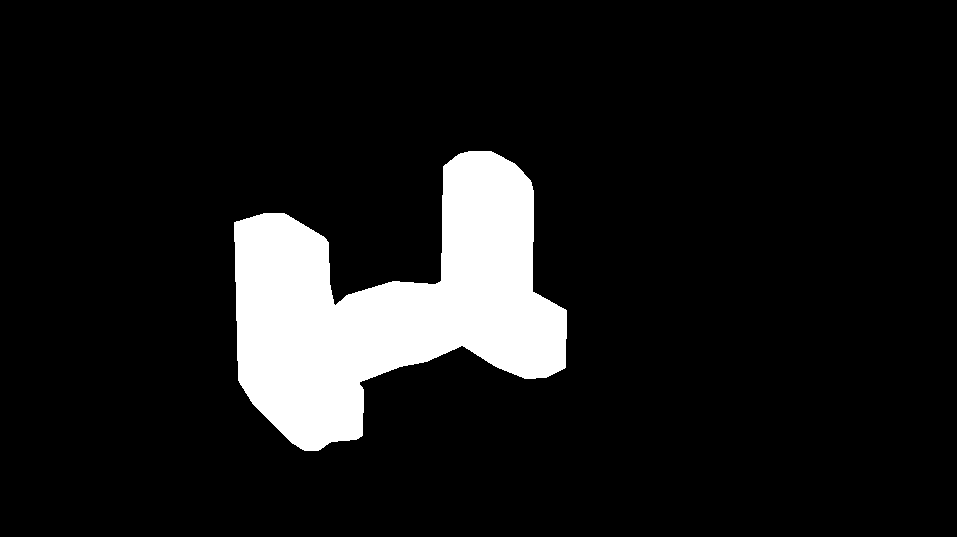}
    \end{subfigure}
    
        \begin{subfigure}[t]{0.23\textwidth} 
        \centering
        \includegraphics[width=\textwidth]{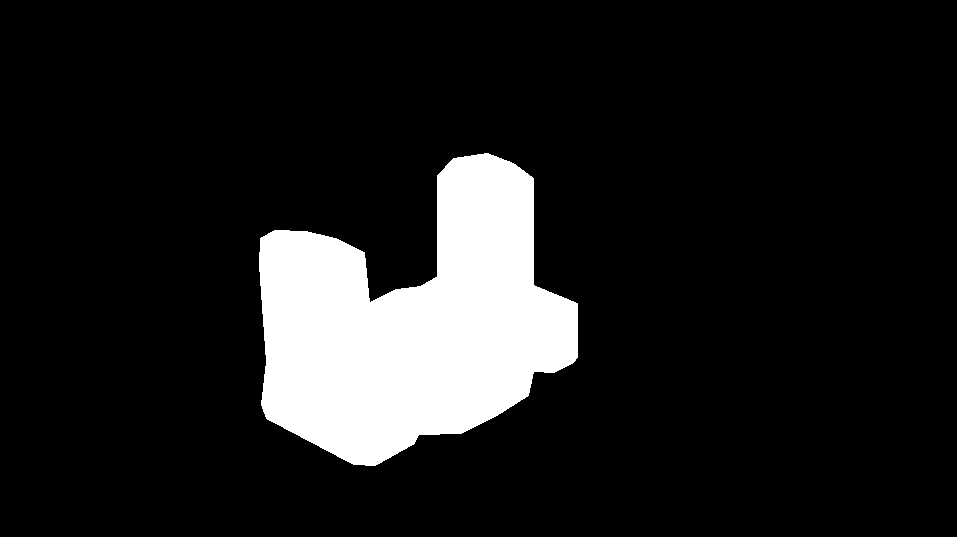}
    \end{subfigure}
        \begin{subfigure}[t]{0.23\textwidth} 
        \centering
        \includegraphics[width=\textwidth]{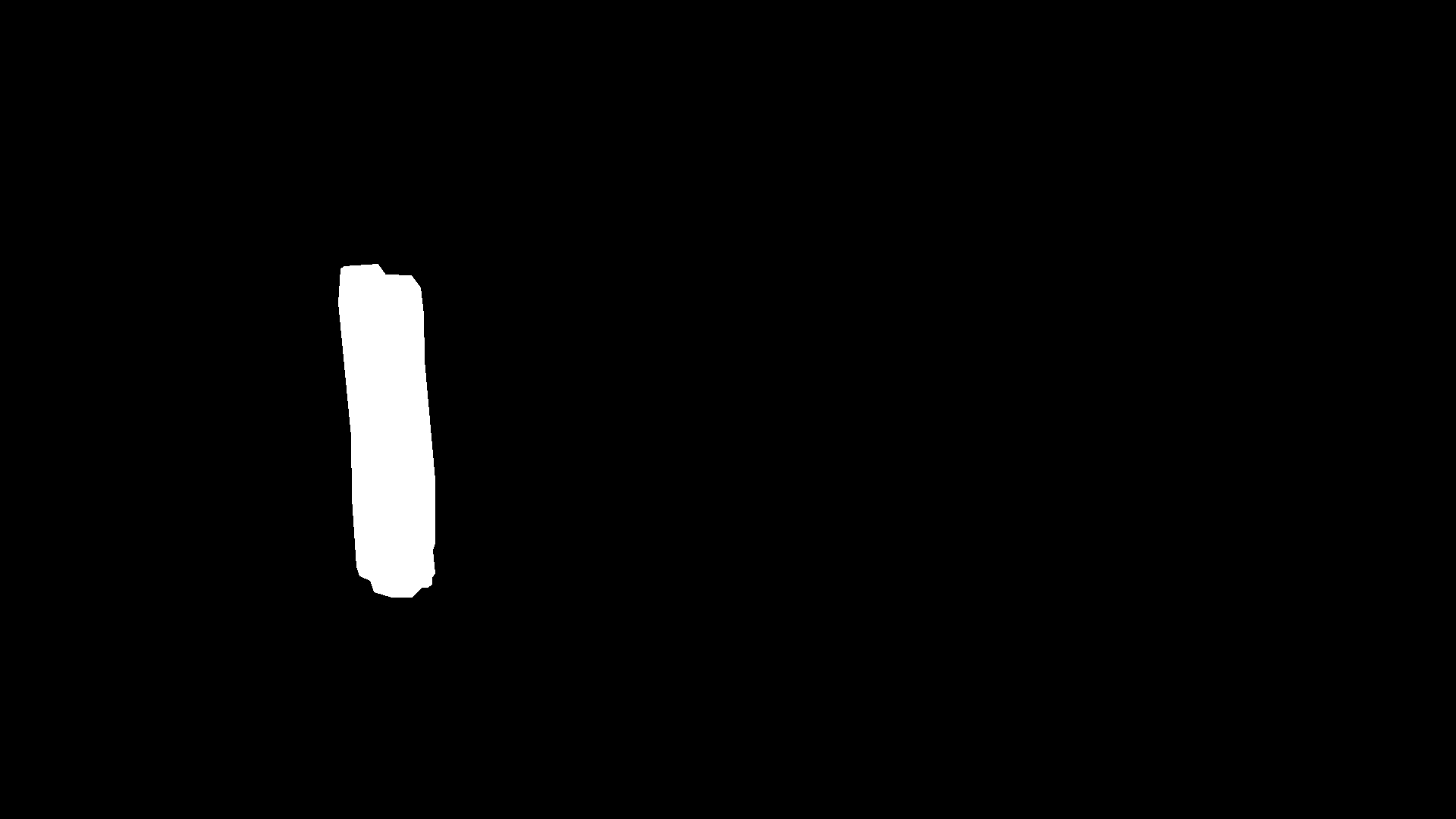}
    \end{subfigure}
    \caption{Example of SAM-2 mask inconsistencies. \textbf{Top Left}: 2D image of the building of interest. \textbf{Top right}: SAM-2 mask of the same frame. \textbf{Bottom-left}: SAM-2 mask in a later frame after some rotation with the parking lot masked in erroneously. \textbf{Bottom-right}: SAM-2 mask in a later frame, with the second tower missing.}
    \label{fig:figure1}
\end{figure}

\subsubsection{Morphological Dilation and Erosion}
The morphological dilation $\oplus$ of an image $I$ by a filter $F$ is defined by

\begin{equation}
    I  \oplus F = \bigcup_{p\in F} I_p,
    \end{equation}    
where $I_p$ denotes the image $I$ shifted by pixel $p=(p_x,p_y)$. Roughly speaking, this grows the image by sliding the filter along the pixels of the original image. This operation can be performed iteratively to further extend the image. At each step, the output image of the previous iteration is dilated again using the same filter. 

Before the morphological dilation, we add an additional optional morphological erosion step (with the same parameters), which is the inverse operation of morphological dilation. That is, morphological erosion shrinks the mask according to the filter. This step serves two purposes. Firstly, depending on the outline of the building, we may not want the dilated mask to extend past the building's true boundaries. Performing the erosion-dilation steps allows the mask to be roughly as tight-fitting as the initial SAM2 mask, while still morphologically filling in holes in the mask. Secondly, at times, SAM2 masks introduce small false-positive pixels away from the central object in the form of noise pixels. This step allows the pipeline to erode away this type of noise.

\subsubsection{Canny Edge Extraction}
Canny edge extraction is a well-known classical algorithm for extracting lines using the image's gradient. The Canny algorithm is well-known in computer vision. As such, we refer readers to the original paper for details \citep{canny}. 


\subsubsection{Ramer–Douglas–Peucker Contour Refinement}
The Ramer–Douglas–Peucker algorithm (RDP) \citep{1972rdp} is an iterative contour simplification algorithm that recursively reduces the number of points in the contour. The algorithm is as follows:

\begin{tcolorbox}[colback=white, colframe=black, sharp corners, boxrule=0.5pt]
\small
\noindent
\textbf{Given:}

\noindent
A curve defined by a sequence of ordered points:
\begin{equation}
P = \{P_1, P_2, \ldots, P_n\}, \quad \text{where } P_i = (x_i, y_i)
\end{equation}
and a threshold parameter:
\begin{equation}
\epsilon > 0
\end{equation}

\noindent
\textbf{Algorithm:}

\begin{enumerate}
    \item Start with endpoints \( P_1 \) and \( P_n \). Define the line segment:
    \[
    \overline{P_1P_n}
    \]

    \item For each point \( P_i = (x_i, y_i) \) with \( i \in [2, n-1] \), compute the perpendicular distance \( d_i \) to the line \( \overline{P_1P_n} \):
\begin{equation}
    d_i = \frac{|(x_n - x_1)(y_1 - y_i) - (x_1 - x_i)(y_n - y_1)|}{\sqrt{(x_n - x_1)^2 + (y_n - y_1)^2}}
\end{equation}

    \item Determine the maximum distance:
\begin{equation}
    d_{\text{max}} = \max_{i=2,\ldots,n-1} d_i
\end{equation}

    \item If \( d_{\text{max}} > \varepsilon \), let \( P_k \) be the point at which the maximum distance occurs. Recursively apply the algorithm to the subsets:
\begin{equation}    
    \{P_1, P_2, \ldots, P_k\} \quad \text{and} \quad \{P_k, P_{k+1}, \ldots, P_n\}
\end{equation}

    \item If \( d_{\text{max}} \leq \varepsilon \), approximate the entire set by the line segment \( \overline{P_1P_n} \), and keep only \( P_1 \) and \( P_n \).
\end{enumerate}

\noindent
\textbf{Output:}

\noindent
A reduced set of points \( P' \subseteq P \), forming a piecewise linear approximation within tolerance \( \epsilon \).

\end{tcolorbox}

We used $\epsilon = \frac{p}{500}$, as the ratio of the perimeter $p$ of the contour being simplified, allowing longer perimeter contours to retain more complexity. The resulting building contours became linear interpolations of approximately 20-40 points. We then filled the contour to create the refined mask, which addresses the aforementioned issues. We tested both our own implementation and an OpenCV implementation and found no discernible difference.

\subsection{2DGS+}
We use a state-of-the-art implementation of Gaussian Splatting built on the foundations of 2DGS. This implementation, which we dubbed 2DGS+ \citep{2dgsp}, is an unpublished fork of the 2DGS repository and combines many recent advances in Gaussian Splatting, improving 2DGS with ideas from AbsGS \citep{absgs}, PixelGS \citep{pixelgs}, TrimGS \citep{trimgs}, AtomGS \citep{atomgs}, GaussianPro \citep{gaussianpro}, and Taming-3DGS \citep{tamings}.

In our experiments, we found that the Gaussian Splatting representations generated from 2DGS+ have fewer floaters (free-floating Gaussians in mid-air), are smoother, and produce better depth maps, resulting in significantly higher-quality meshes during TSDF fusion meshing (see section \ref{results:mesh_extraction}). We tuned the hyperparameters for 3D mesh extraction quality (not novel view synthesis quality). We used the progressive training modality from GaussianPro, densifying the Gaussians until the 25,000th iteration. We used $\lambda_{depth} = 0.2$ and $\lambda_{normal} = 0.1$.

\subsection{TSDF Fusion and Mesh extraction}
Like GS2Mesh and 2DGS, we used the Open3D implementation of the TSDF Fusion algorithm to convert depth maps into colored 3D meshes. We adapted the 2DGS mesh extraction module, adding a depth map smoothing step and tuning the hyperparameters. We used the 2DGS+ implementation \citep{20242dgs} of 2DGS with its bounded mesh extraction module.

In early experiments, we noticed the presence of non-smoothness in the depth maps of certain complex buildings, which resulted in poor-quality mesh reconstructions. In these cases, applying a simple Gaussian blur to the depth maps before TSDF fusion greatly improved the mesh reconstruction quality.

\begin{figure}[h!]
    \begin{subfigure}[t]{0.23\textwidth} 
        \centering
        \includegraphics[width=\textwidth]{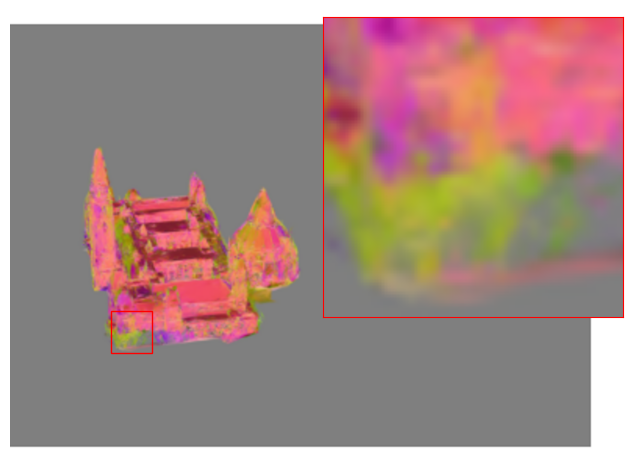}
    \end{subfigure}
    \begin{subfigure}[t]{0.23\textwidth} 
        \centering
        \includegraphics[width=\textwidth]{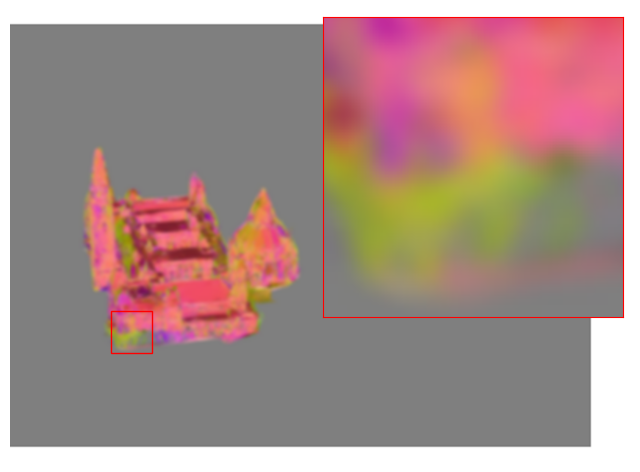}
    \end{subfigure}

    \begin{subfigure}[t]{0.23\textwidth} 
        \centering
        \includegraphics[width=\textwidth]{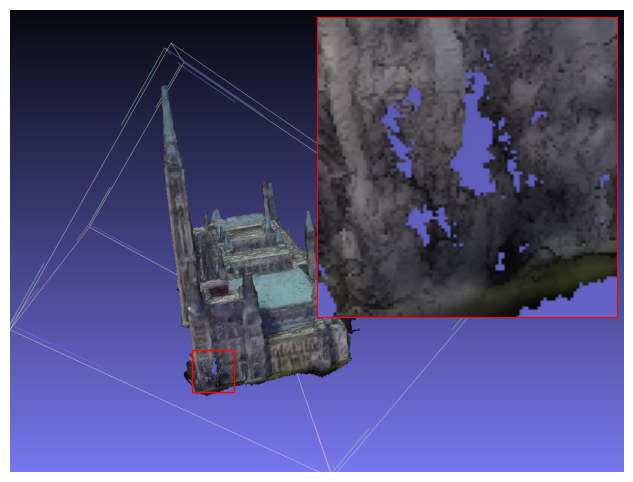}
    \end{subfigure}
    \begin{subfigure}[t]{0.23\textwidth} 
        \centering
    \includegraphics[width=\textwidth]{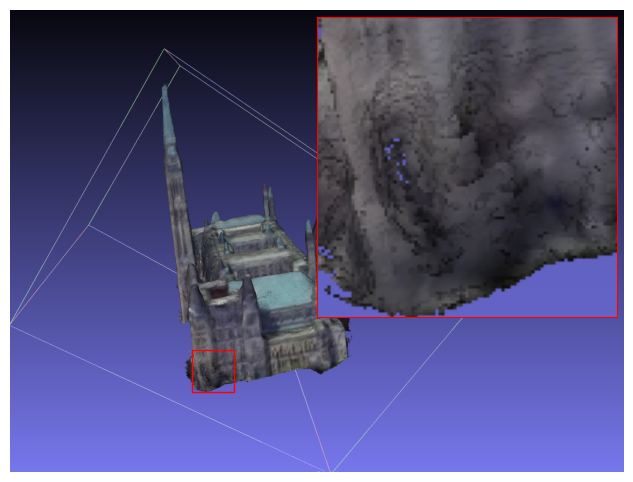}
    \end{subfigure}

    \caption{Example of depth blurring. \textbf{Top-left}: Raw depth map. \textbf{Top-right}: Smoothed depth map. \textbf{Bottom-left}: 3D Mesh from raw depth map. \textbf{Bottom-right}: 3D Mesh from the smoothed depth map. }
    \label{fig:depth_blur}
\end{figure}

After mesh extraction, like GS2Mesh and 2DGS, we used Open3D to clean the mesh by computing mesh clusters, removing unreferenced vertices, and degenerate triangles. Unlike GS2Mesh and 2DGS, we discarded all but the largest mesh cluster.
\section{Results}
Although there are multiple Gaussian Splatting-based methods for 3D mesh extraction of entire scenes, we found GS2Mesh to be the only method comparable in its application: mesh extraction for single objects from text-based or click-based prompts with optional automation. Additionally, we used it as a starting point to develop our framework. As such, we focus our comparisons against it.

\begin{figure}[h]
\centering
\includegraphics[width=0.5\textwidth]{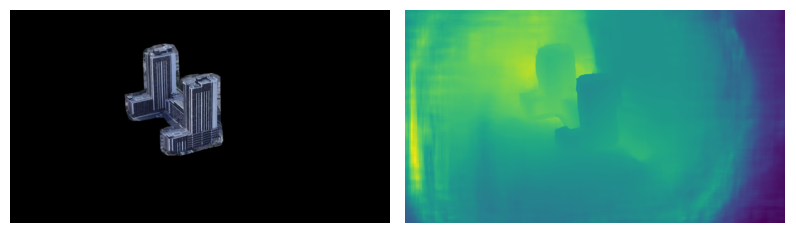}
\caption{Example of DLNR failure case. \textbf{Left}: 2D image of the building of interest with background masked out. \textbf{Right}: DLNR stereo depth reconstruction from building with background masked out. }\label{img:dlnr}
\end{figure}
\subsection{Preliminary Experiments}
Our preliminary experiments on 3D building mesh extraction from remote sensing images showed that the GS2Mesh 3D reconstruction pipeline, which uses DLNR \citep{2023DLNR}, a pretrained neural-network-based stereo depth reconstruction and depth map fusion, only matched the quality of 2DGS depth map fusion in the best-case scenario. More often than not, it struggled in the 3D reconstruction of buildings due to two key issues: poor mask quality and poor stereo depth estimation. DLNR occasionally struggled with stereo depth reconstruction, likely because it was not trained on a remote sensing image dataset and had difficulty interpreting large-scale aerial scenes. This issue becomes particularly pronounced when stereo depth extraction is applied after masking, as the neural network loses the context of the background. In Figure \ref{img:dlnr}, we observe that DLNR fails completely when performing depth extraction on a building with the background masked out. This motivated us to seek alternative solutions for the 3DGS-DLNR pipeline components in GS2Mesh.

Additional preliminary experiments with 2DGS+ showed that depth maps of certain complex buildings exhibited non-smoothness, resulting in poor-quality 3D meshes. We addressed this issue with the simple solution of smoothing the depth map using a Gaussian filter. An example is shown in Figure \ref{fig:depth_blur}. After applying the filter, the holes in the zoomed-in region were almost completely filled. This particular issue arose due to poor visual coverage of the zoomed-in region in the initial training data.

\begin{figure*}[htbp]
    \begin{subfigure}[t]{0.31\textwidth} 
        \centering
        \includegraphics[width=\textwidth]{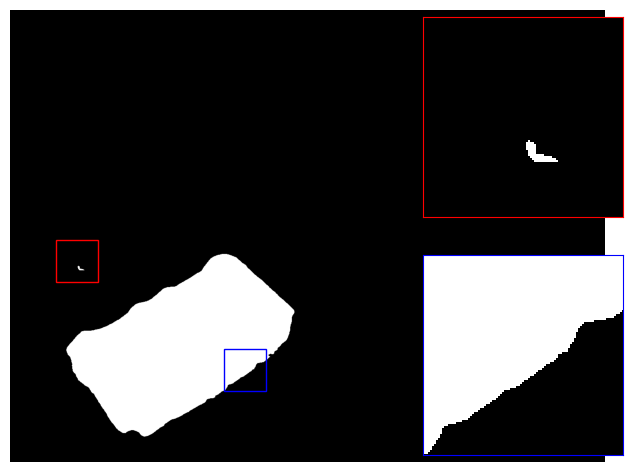}
    \end{subfigure}
        \begin{subfigure}[t]{0.31\textwidth} 
        \centering
        \includegraphics[width=\textwidth]{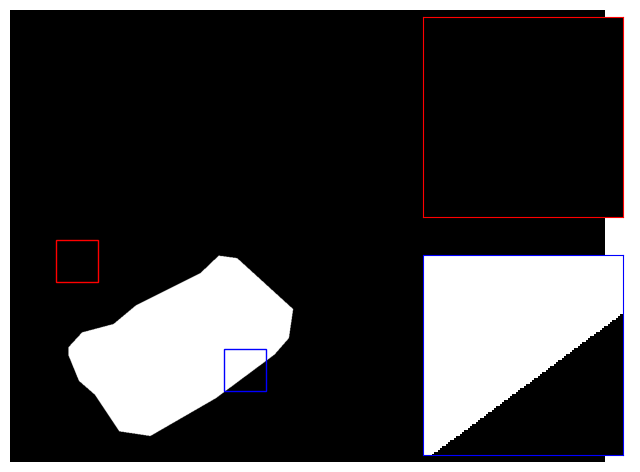}
    \end{subfigure}
    \begin{subfigure}[t]{0.31\textwidth} 
        \centering
        \includegraphics[width=\textwidth]{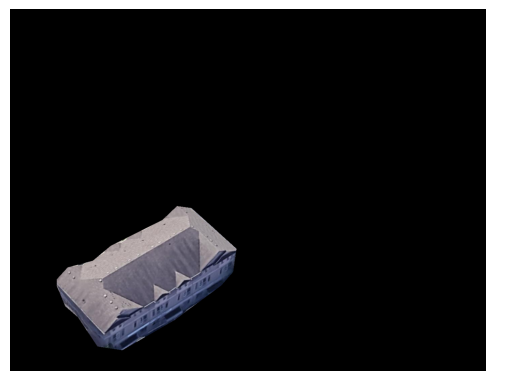}
    \end{subfigure}

        \begin{subfigure}[t]{0.31\textwidth} 
        \centering
        \includegraphics[width=\textwidth]{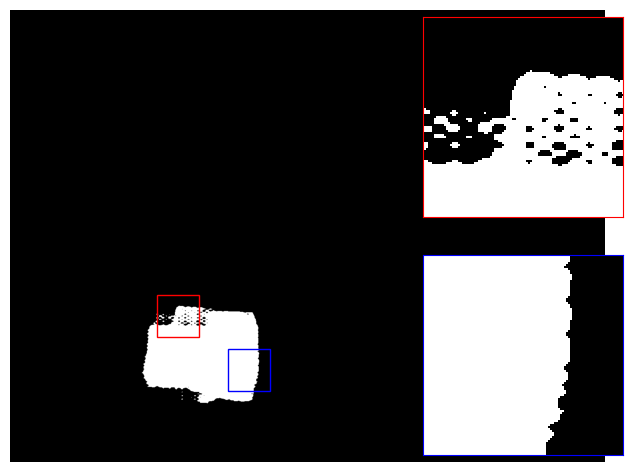}
    \end{subfigure}
        \begin{subfigure}[t]{0.31\textwidth} 
        \centering
        \includegraphics[width=\textwidth]{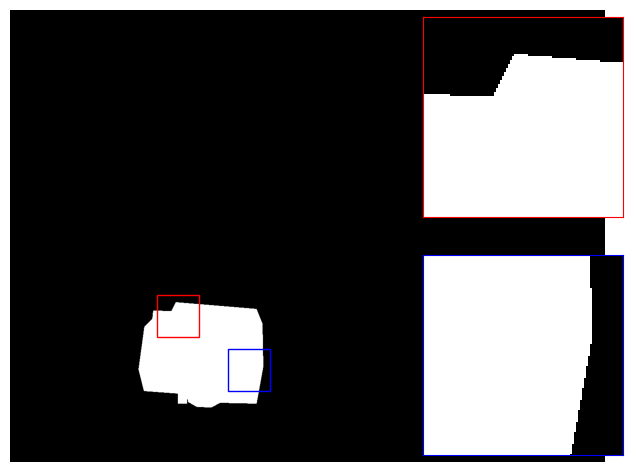}
    \end{subfigure}
    \begin{subfigure}[t]{0.31\textwidth} 
        \centering
        \includegraphics[width=\textwidth]{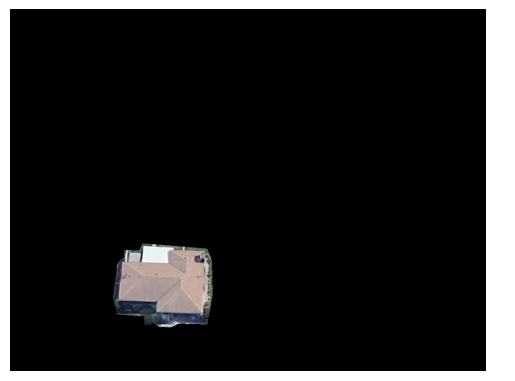}
    \end{subfigure}
    \caption{Example of mask refinement with noise removal \textbf{(top)} and hole filling \textbf{(bottom)}: \textbf{Left}: Raw SAM-2 Mask. \textbf{Middle}: Refined Mask. \textbf{Right}: masked Image. Regions of interest have been highlighted. Red indicates a region with false positives or false negative pixels, blue indicates a region where the contour was refined and straightened.}
    \label{fig:maskref1}
\end{figure*}

\subsection{Mask Refinement} \label{results:mask_refinement}
SAM2 masks sometimes included false-positive pixels away from the object of interest. This issue is visible in the left-most subfigure of Figure \ref{fig:maskref1} (top row). This is likely caused by false-positive segmentation due to the presence of a neighboring building's rooftop with a similar shape and color in the original full image. Our mask refinement algorithm successfully removed the false-positive pixels and refined the mask contour.

SAM2 also occasionally produced masks with holes or false negatives. We observed that this occurred more frequently when part of the building exhibited significantly different coloration from the rest of the structure. An example is shown in Figure \ref{fig:maskref1} (bottom row), where the garage roof has a different color. Our mask refinement algorithm was able to complete the mask and refine the mask contour.
\begin{table*}[htbp]
\centering
\caption{Comparison 2D Novel View Synthesis Metrics Between Gaussian Splatting Modules}
\begin{tabular}{l|c c c| c c c}
\hline
       & \multicolumn{3}{c|}{2DGS+ (in Our Pipeline)}        & \multicolumn{3}{c}{3DGS (in GS2Mesh)}     \\
Building                            & SSIM $\uparrow$   & PSNR $\uparrow$   & LPIPS $\downarrow$  & SSIM $\uparrow$   & PSNR $\uparrow$   & LPIPS $\downarrow$ \\ \hline
Icon                      & 0.9944          & 37.61         & 0.0100          & 0.9886          & 37.23         & 0.0162          \\ 
Canada Parliament         & 0.9894          & 35.11         & 0.0156          & 0.9321          & 29.75         & 0.1076          \\ 
CN Tower                  & 0.9977          & 40.01         & 0.0039          & 0.9946          & 42.63         & 0.0092          \\ 
Laurel Heights            & 0.9948          & 38.13         & 0.0089          & 0.9895          & 38.04         & 0.0127          \\ 
Perimeter Institute       & 0.9950          & 39.40         & 0.0080          & 0.9935          & 43.93         & 0.0149          \\
Dana Porter Library & 0.9746 & 32.25 & 0.0330 & 0.9923& 38.40 & 0.0109 \\
Townhouse & 0.9946 & 36.82 &0.0087 & 0.9919 & 39.66 & 0.0118 \\ \hline
Average & \textbf{0.9915} & 37.05 & \textbf{0.0127} & 0.9833 & \textbf{38.52} & 0.0260 \\ \hline

\end{tabular}
\label{tab:2Dscores}
\end{table*}
\begin{table*}[htbp]
\centering
\caption{Comparison of 3D-SSIM Scores Between Our Pipeline and GS2Mesh}
\begin{tabular}{l|c|c|c|c}
\hline
                        & \multicolumn{2}{c|}{GBM (Ours) 3D-SSIM $\uparrow$} & \multicolumn{2}{c}{GS2Mesh 3D-SSIM  $\uparrow$} \\ 
Building                          & \multicolumn{1}{c}{Average (31 frames)} &\multicolumn{1}{c|}{Minimum}         & \multicolumn{1}{c}{Average (31 frames)} &\multicolumn{1}{c}{Minimum}          \\ \hline
Icon                      & 0.9287           & 0.9158                    & 0.8355            & 0.8000                    \\ 
Canada Parliament         & 0.8535           & 0.8458                    & 0.1462            & 0.0703                   \\ 
CN Tower                  & 0.9841           & 0.9771                    & 0.9397            & 0.9282                   \\ 
Laurel Heights            & 0.9566           & 0.9529                    & 0.9427            & 0.9412                   \\
Perimeter Institute       & 0.9747           & 0.9726                    & 0.9475            & 0.9431                   \\ 
Data Porter Library       & 0.8214           & 0.8078                    & 0.7054            & 0.6862                   \\
Townhouse                 & 0.9780            & 0.9749                    & 0.9785           & 0.9761                   \\ \hline
Average & \textbf{0.9281} & \textbf{0.9201} & 0.7850 & 0.7536 \\ \hline
\end{tabular}
\label{tab:3d_ssim_comparison}
\end{table*}
 
\subsection{View Synthesis}
Compared to GS2Mesh, we replaced the Gaussian Splatting module in the pipeline, transitioning from the original implementation of 3DGS to 2DGS+. As such, we compare the training set novel view synthesis results of 3DGS and 2DGS+ across our 7 scenes. We did not use the standard MipNeRF-360 convention of a training/testing split (leaving out one out of every eight images). Instead, we trained on the entirety of the Google Earth Studio footage, as we wanted smooth coverage of the 360$\degree$ camera rotation for 3D reconstruction purposes. The novel view synthesis scores on the training set are provided in Table \ref{tab:2Dscores}.

\begin{figure*}[htbp]
    \begin{subfigure}[t]{0.245\textwidth} 
        \centering
        \includegraphics[width=\textwidth]{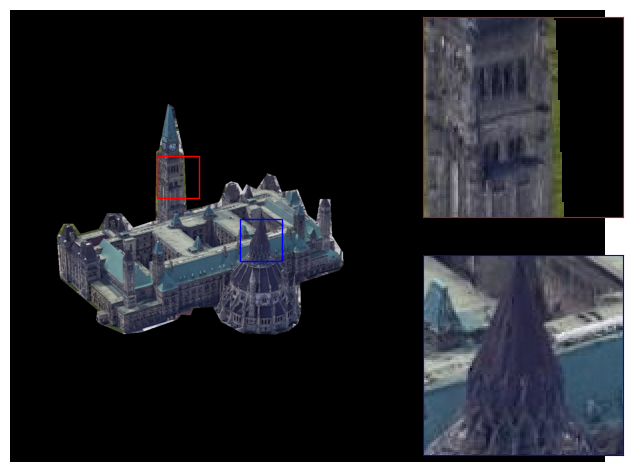}
    \end{subfigure}
        \begin{subfigure}[t]{0.245\textwidth} 
        \centering
        \includegraphics[width=\textwidth]{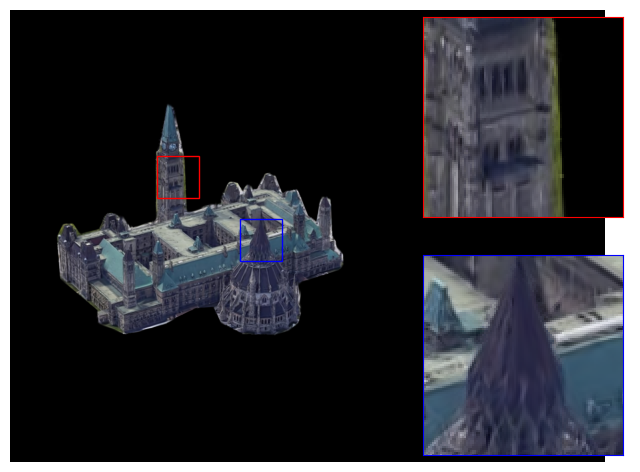}
    \end{subfigure}
    \begin{subfigure}[t]{0.245\textwidth} 
        \centering
        \includegraphics[width=\textwidth]{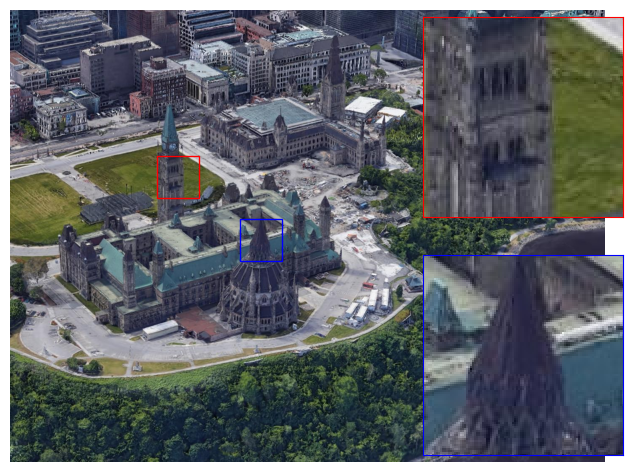}
    \end{subfigure}
    \begin{subfigure}[t]{0.245\textwidth} 
        \centering
        \includegraphics[width=\textwidth]{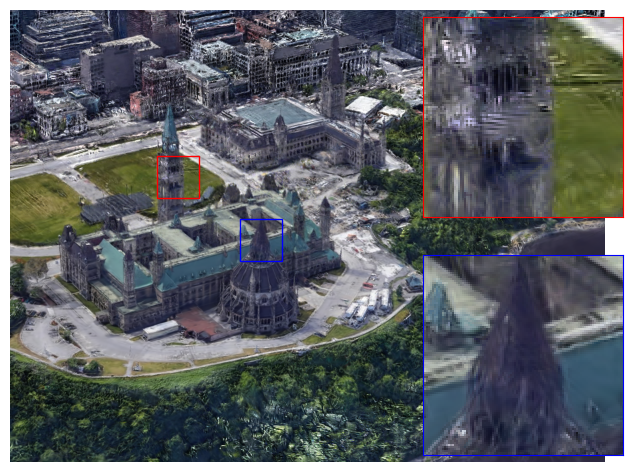}
    \end{subfigure}

    \begin{subfigure}[t]{0.245\textwidth} 
        \centering
        \includegraphics[width=\textwidth]{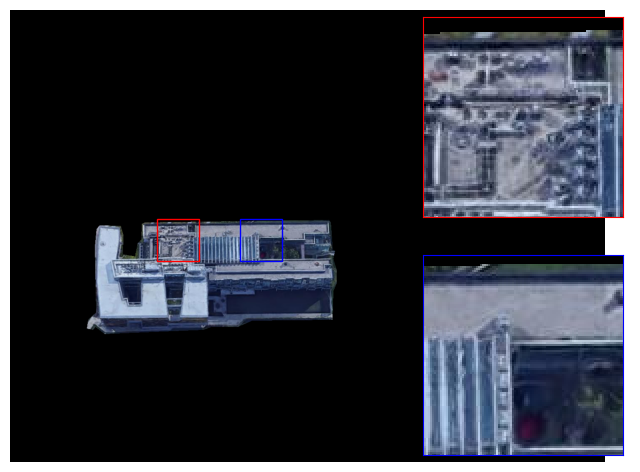}
    \end{subfigure}
        \begin{subfigure}[t]{0.245\textwidth} 
        \centering
        \includegraphics[width=\textwidth]{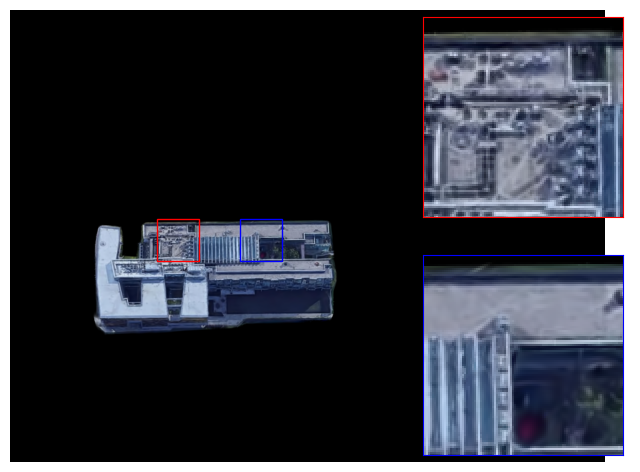}
    \end{subfigure}
    \begin{subfigure}[t]{0.245\textwidth} 
        \centering
        \includegraphics[width=\textwidth]{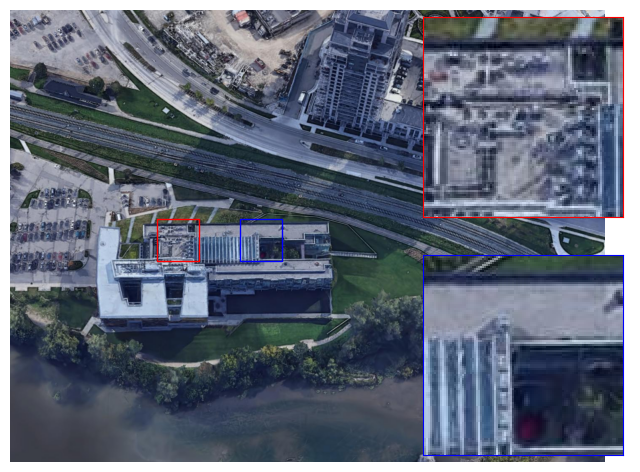}
    \end{subfigure}
    \begin{subfigure}[t]{0.245\textwidth} 
        \centering
        \includegraphics[width=\textwidth]{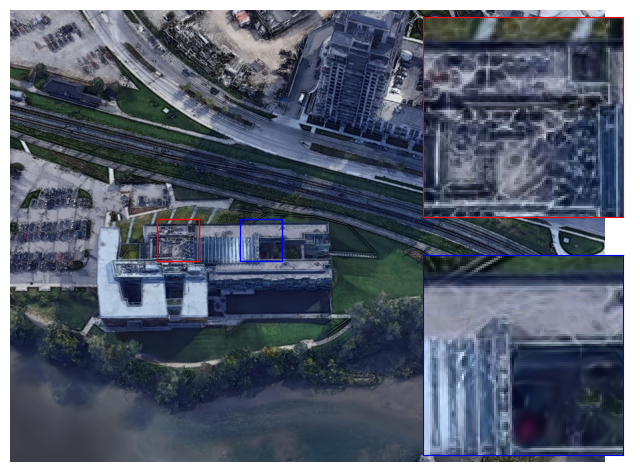}
    \end{subfigure}

        \begin{subfigure}[t]{0.245\textwidth} 
        \centering
        \includegraphics[width=\textwidth]{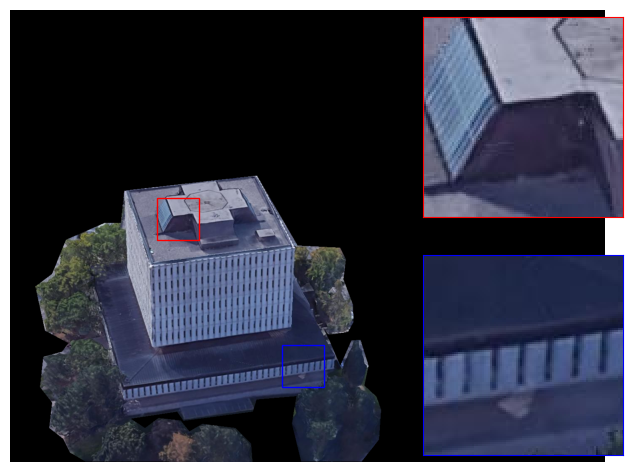}
    \end{subfigure}
        \begin{subfigure}[t]{0.245\textwidth} 
        \centering
        \includegraphics[width=\textwidth]{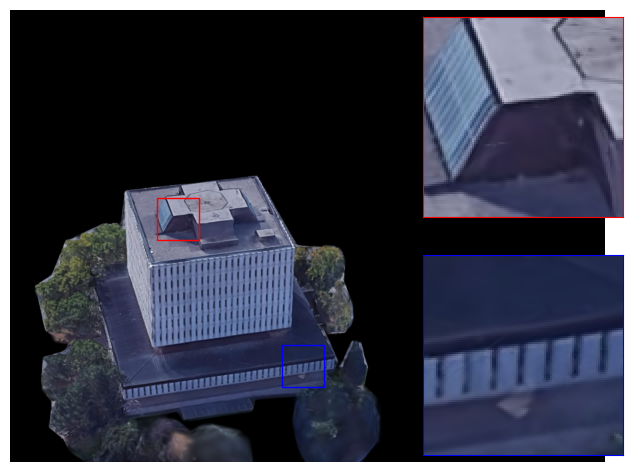}
    \end{subfigure}
    \begin{subfigure}[t]{0.245\textwidth} 
        \centering
        \includegraphics[width=\textwidth]{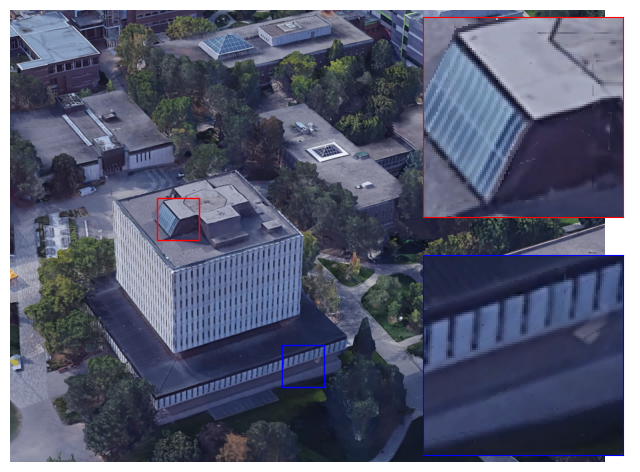}
    \end{subfigure}
    \begin{subfigure}[t]{0.245\textwidth} 
        \centering
        \includegraphics[width=\textwidth]{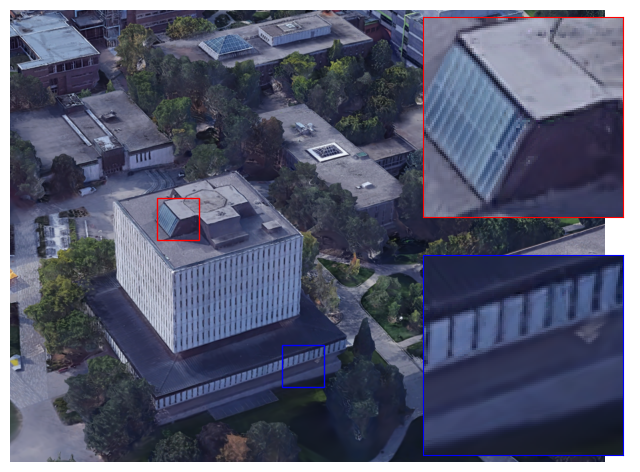}
    \end{subfigure}

    \caption{Visualization of 2D novel view synthesis results. \textbf{Left to Right: 2DGS Ground Truth; 2DGS Sythesized Image; 3DGS Ground Truth; 3DGS Synthesized Image}. We note that for all cases including those not shown, the results were nearly visually indistinguishable from ground truth by human eyes, except for 3DGS trained on the Canada Parliament scene (top right).}
    \label{fig:view_synthesis}
\end{figure*}
For 2DGS+ in our pipeline, we trained the model on the masked building without the background. The model's synthesized images were then compared to images of the building with the background masked out. On the other hand, in the GS2Mesh pipeline, the 3DGS training occurs before masking and is conducted on the entire image, including the background. Its 3DGS module generates images with the entire background, and the results are compared with unmasked ground truth images.

On average, we found that 2DGS+ produced higher SSIM and LPIPS scores, whereas 3DGS achieved higher PSNR scores. We note that SSIM and LPIPS are typically better indicators of human visual perception than PSNR. Nonetheless, both models achieved novel view synthesis results that were nearly indistinguishable from ground truth images, with the exception of 3DGS struggling on the Canada Parliament scene, as shown in the top row of Figure \ref{fig:view_synthesis} (see the central tower highlighted in the top-right zoom-in). This is also reflected in Table \ref{tab:2Dscores}, where the 3DGS scores for the Canada Parliament scene are significantly lower than for other scenes.

\subsection{3D Building Mesh Extraction} \label{results:mesh_extraction}
\begin{figure*}[htbp]
    \begin{subfigure}[t]{0.19\textwidth} 
        \centering
    \includegraphics[width=\textwidth]{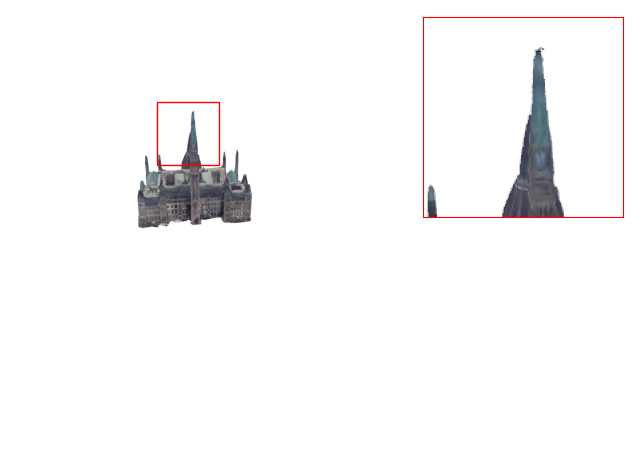}
    \end{subfigure}
    \begin{subfigure}[t]{0.19\textwidth} 
        \centering
        \includegraphics[width=\textwidth]{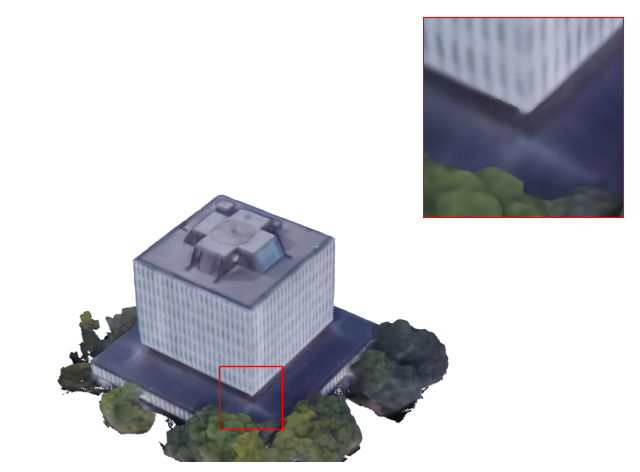}
    \end{subfigure}
        \begin{subfigure}[t]{0.19\textwidth} 
        \centering
        \includegraphics[width=\textwidth]{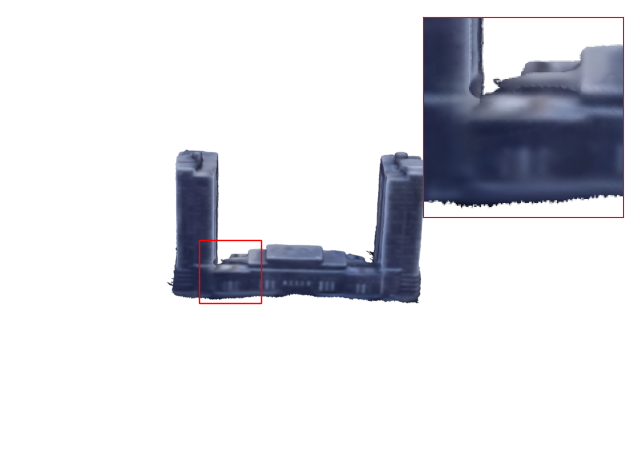}
    \end{subfigure}
    \begin{subfigure}[t]{0.19\textwidth} 
        \centering
        \includegraphics[width=\textwidth]{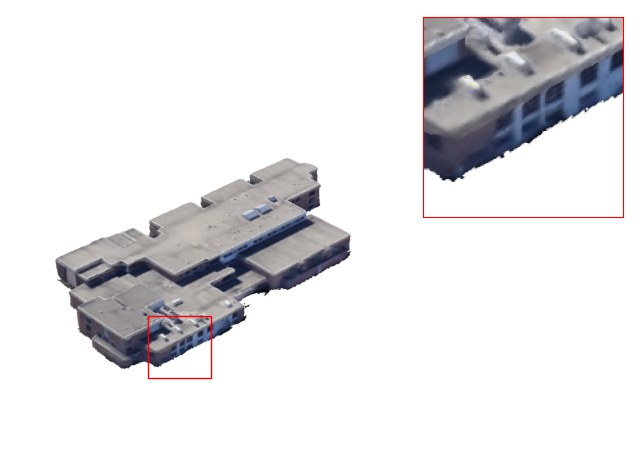}
    \end{subfigure}
    \begin{subfigure}[t]{0.19\textwidth} 
        \centering
        \includegraphics[width=\textwidth]{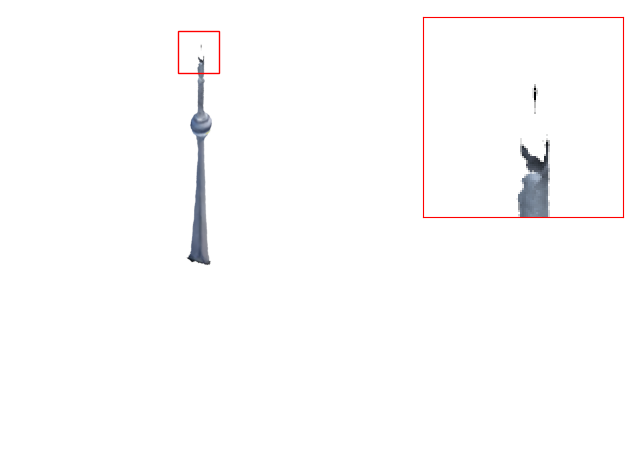}
    \end{subfigure}

    \begin{subfigure}[t]{0.19\textwidth} 
        \centering
        \includegraphics[width=0.9\textwidth]{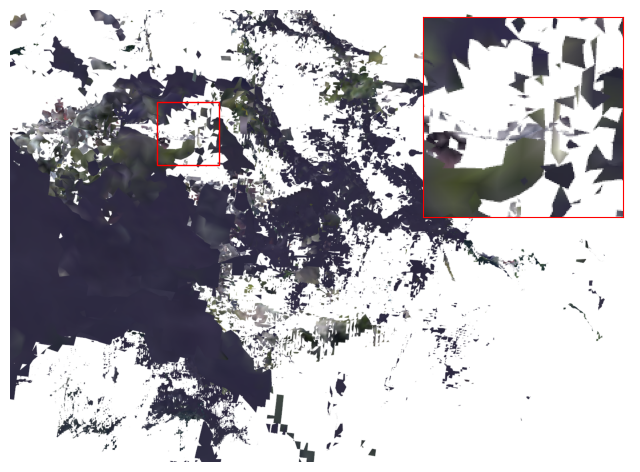}
    \end{subfigure}
    \begin{subfigure}[t]{0.19\textwidth} 
        \centering
        \includegraphics[width=\textwidth]{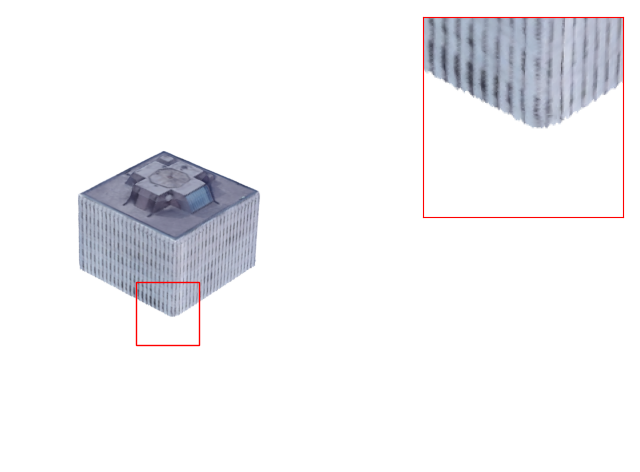}
    \end{subfigure}
        \begin{subfigure}[t]{0.19\textwidth} 
        \centering
        \includegraphics[width=\textwidth]{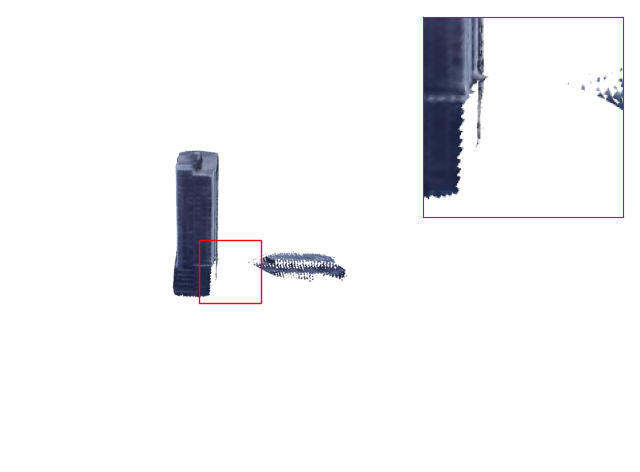}
    \end{subfigure}
    \begin{subfigure}[t]{0.19\textwidth} 
        \centering
        \includegraphics[width=\textwidth]{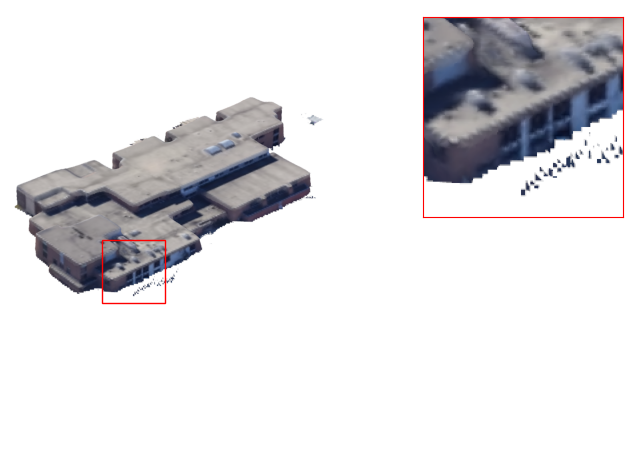}
    \end{subfigure}
    \begin{subfigure}[t]{0.19\textwidth} 
        \centering
        \includegraphics[width=\textwidth]{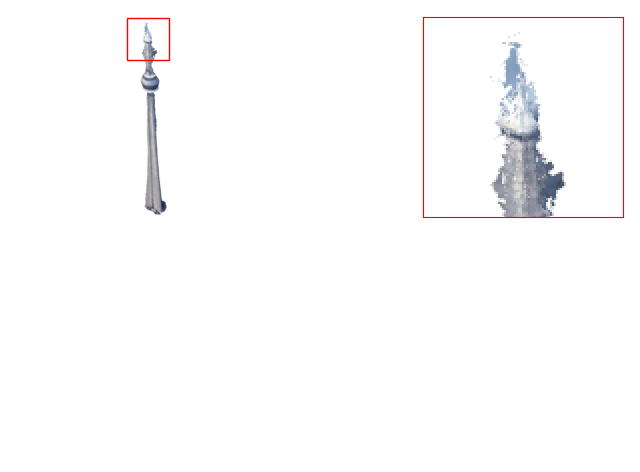}
    \end{subfigure}

    \caption{Visualization of 3D colored mesh results, showcasing imperfections in both methods. \textbf{Left to Right}: Canada Parliament; Dana Porter Library; ICON; Laurel Heights; CN Tower. \textbf{Top row} are our results, \textbf{Bottom row} are GS2Mesh results.}
    \label{fig:3DResults}
\end{figure*}
As shown in Table \ref{tab:3d_ssim_comparison}, our building extraction results are superior to GS2Mesh, with the exception of the Townhouse scene, where both methods scored similarly, within $\sim$0.0005 3D-SSIM. The results are somewhat skewed by the Canada Parliament scene (first column of Figure \ref{fig:3DResults}), where GS2Mesh 3D mesh reconstruction failed completely due to the significant presence of floaters during the training of its 3DGS module. This failure caused its DLNR depth reconstruction module to produce completely inaccurate depth maps. The poor mesh quality resulted in extremely low 3D-SSIM scores, as reflected in Table \ref{tab:3d_ssim_comparison}. The resulting mesh was completely unrecognizable. The poor performance of the 3DGS module of GS2Mesh for this scene is also reflected in its PSNR score in Table \ref{tab:2Dscores}, which is approximately 10 dB lower than for other scenes. GS2Mesh also partially failed on the ICON scene and Dana Porter Library scene, where the lack of user re-prompting caused the masking to be inconsistent across the initial training images, resulting in missing geometric features and incomplete meshing.

Our method produced less noisy meshes overall and cleaner mesh boundaries. This improvement is reflected in the 3D-SSIM scores and can be seen in the fourth column of Figure \ref{fig:3DResults}. However, both GS2Mesh and our method struggled with sharp structural elements, such as the top of the CN Tower, as shown in the fifth column of Figure \ref{fig:3DResults}.

\subsection{Ablation Studies} \label{results:ablation}
\begin{table*}[ht]
\centering
\caption{Ablation Study on 2DGS+ for View Synthesis and 3D Meshing}
\label{tab:ablation}
\begin{adjustbox}{max width=\textwidth}
\begin{tabular}{l |r r r| r r r| r r| c}
\hline
 & \multicolumn{3}{c|}{Train} & \multicolumn{3}{c|}{Test} & \multicolumn{2}{c|}{3DSSIM$\uparrow$} & Training Time$\downarrow$ \\
 & PSNR$\uparrow$ & SSIM$\uparrow$ & LPIPS$\downarrow$ & PSNR$\uparrow$ & SSIM$\uparrow$ & LPIPS$\downarrow$ & Mean & Min &  \\
\hline
2DGS & 34.0 & .987 & .021 & 28.5 & .954 & .045 & .876 & .865 & 0:11:24 \\
2DGS+Trim & 36.4 & .992 & .011 & 28.5 & .955 & .042 & .882 & .877 & 0:13:05 \\
2DGS+Trim+Prog-Norm & 36.7 & .992 & .013 & 28.4 & .952 & .043 & .881 & .876 & 0:12:03 \\
2DGS+Trim+Prog+Norm & 36.8 & .992 & .013 & 28.3 & .951 & .044 & .884 & .870 & 0:11:49 \\
2DGS+Trim+Prog+EANorm* & 36.6 & .992 & .013 & 28.4 & .952 & .042 & .891 & .887 & 0:15:10 \\
2DGS+Trim+Fast+Norm & 36.8 & .993 & .011 & 28.7 & .955 & .041 & .882 & .877 & 0:14:34 \\
2DGS+Trim+Prog+EANorm+Appearance & 36.7 & .992 & .013 & 28.5 & .952 & .043 & .884 & .879 & 0:15:21 \\
2DGS+Trim+Prog+EANorm+Appearance+60k & 38.2 & .994 & .010 & 28.2 & .949 & .045 & .867 & .863 & 0:30:26 \\
\hline
\end{tabular}
\end{adjustbox}
\end{table*} 
The ablation study is performed on a separate Westminster Palace scene comparing the improvements of 2DGS+ to 2DGS as a baseline. A mip-NeRF \citep{mipnerf} style training/testing split was used. Both view synthesis and 3D mesh quality assessments are used. Additionally, training time was measured. Trim-Gaussian's \citep{trimgs} and PixelGS's \citep{pixelgs} contribution-based Gaussian pruning is denoted as Trim. Progressive propagation from GaussianPro\citep{gaussianpro} is denoted as Prog. The standard 2DGS normal-based supervision is denoted as Norm. The Edge-Aware Normal/Geometry Supervision from AtomGS \citep{atomgs} is denoted EANorm. The Appearance Network from GOF \citep{2024GOF} is denoted as Appearance. Fast-SSIM from Taming-GS \citep{tamings} is denoted as Fast. Doubled training iteration is denoted as 60k. 

\subsection{Discussions}
In terms of experiments, for the Dana Porter Library scene (second column of Figure \ref{fig:3DResults}), without user re-prompting, the base of the building was masked inconsistently due to the presence of occluding trees, which resulted in messy GS2Mesh reconstruction. Instead, for GS2Mesh training on this scene, we found that prompting for the large cubical structure without the first floor and the trees resulted in better reconstruction. Our pipeline was able to properly capture the building base along with the surrounding trees. For the ICON scene (third column of Figure \ref{fig:3DResults}), without re-prompting, the mask for the second tower and part of the connected structure was missing in some frames, resulting in an incomplete reconstruction for GS2Mesh. We note that both methods struggled with sharp, needle-like building structure elements, such as the top of the CN Tower. This was likely due to a combination of the difficulty in masking needle-like objects and the challenge of producing depth maps for these objects.

The GS2Mesh failure on the Canada Parliament scene was unexpected. We believe the failure occurred in the 3DGS module, as we noticed a significant amount of floaters (Gaussians floating in empty space). We did not attempt further fine-tuning of the 3DGS training. These floaters were absent during the 2DGS+ stage in our pipeline, even without additional fine-tuning.

We believe further improvements can be made to the mesh boundary at the building base. Although our method's mesh boundaries improved, we still noticed irregularities and poor-quality meshing at the base of buildings. We suspect that this issue, along with the problem of meshing sharp structural elements, can be addressed by further improving building masks—a potential extension to this research. Our improvements to SAM2 masking can also be used independently of the meshing pipeline in other applications of SAM2/GroundedSAM. This topic could constitute a separate line of research apart from Gaussian Splatting-based mesh extraction.

\section{Conclusion}
The ability to generate high-quality 3D building models from minimal user input addresses a key need in digital building monitoring workflows. We developed a robust pipeline capable of automatically extracting a colored 3D mesh of a building using its address, postal code, geographic coordinates, or location name directly. Our pipeline does not require image, video, or point cloud data from the user, and instead extracts the data from Google Earth Studio. We benchmarked our method against GS2Mesh, which, to the best of our knowledge, is the only comparable segmentation-based Gaussian Splatting meshing algorithm publicly available. Our results demonstrate that our approach produces significantly higher-quality meshes and is significantly less prone to failure, allowing for automated extraction of the 3D mesh of any user-designated buildings given Google Earth image coverage. These capabilities open the door to scalable applications in digital twin creation, construction monitoring and automated inspection, disaster monitoring, and many other downstream tasks.



%
\section*{Declaration of competing interest}
The authors declare that they have no known competing financial interests or personal relationships that could have appeared to influence the work reported in this paper.

\section*{Acknowledgements}
This work was supported in part by the Natural Sciences and Engineering Research Council of Canada (NSERC) Discovery Grant (No. RGPIN-2022-03741).

We would like to thank the authors of 2DGS \citep{20242dgs} and GS2Mesh \citep{2024gs2mesh}, and the GitHub contributors of 2DGS+ \citep{2dgsp} for making their codes publicly available.

\bibliographystyle{cas-model2-names}
\bibliography{References}

\newpage
 




\vfill

\end{document}